\documentclass[10pt,twocolumn]{article}

\usepackage[utf8]{inputenc}
\usepackage[T1]{fontenc}
\usepackage{amsmath,amssymb,amsfonts}
\usepackage{graphicx}
\usepackage{booktabs}
\usepackage{hyperref}
\usepackage{xcolor}
\usepackage{float}
\usepackage{caption}
\usepackage[margin=1in]{geometry}
\usepackage{authblk}
\usepackage{enumitem}
\usepackage{times}

\hypersetup{
    colorlinks=true,
    linkcolor=blue,
    citecolor=blue,
    urlcolor=blue
}

\title{\textbf{Beyond Human Performance: A Vision-Language Multi-Agent Approach for Quality Control in Pharmaceutical Manufacturing}}

\author[1]{Subhra Jyoti Mandal}
\author[2]{Lara Rachidi}
\author[2]{Puneet Jain}
\author[1]{Matthieu Duvinage}
\author[1]{Sander W. Timmer}

\affil[1]{GSK, Enterprise AI}
\affil[2]{Databricks}

\date{}

\begin{document}

\maketitle

\begin{abstract}
Colony-forming unit (CFU) detection is critical in pharmaceutical manufacturing, serving as a key component of Environmental Monitoring programs and ensuring compliance with stringent quality standards. Manual counting is labor-intensive and error-prone, while deep learning (DL) approaches, though accurate, remain vulnerable to sample quality variations and artifacts.

Building on our earlier CNN-based framework (Beznik et al., 2020), we evaluated YOLOv5, YOLOv7, and YOLOv8 for CFU detection; however, these achieved only 97.08\% accuracy---insufficient for pharmaceutical-grade requirements. A custom Detectron2 model trained on GSK's dataset of 50,000+ Petri dish images achieved 99\% detection rate with 2\% false positives and 0.6\% false negatives.

Despite high validation accuracy, Detectron2 performance degrades on outlier cases including contaminated plates, plastic artifacts, or poor optical clarity. To address this, we developed a multi-agent framework combining DL with vision-language models (VLMs). The VLM agent first classifies plates as valid or invalid. For valid samples, both DL and VLM agents independently estimate colony counts. When predictions align within 5\%, results are automatically recorded in Postgres and SAP; otherwise, samples are routed for expert review. Expert feedback enables continuous retraining and self-improvement.

Initial DL-based automation reduced human verification by 50\% across vaccine manufacturing sites. With VLM integration, this increased to 85\%, delivering significant operational savings. The proposed system provides a scalable, auditable, and regulation-ready solution for microbiological quality control, advancing automation in biopharmaceutical production.
\end{abstract}


\section{Introduction}

Colony-forming unit (CFU) detection constitutes a foundational process in microbiological quality control, ensuring sterility and microbial safety across pharmaceutical and vaccine manufacturing pipelines. Accurate enumeration of microbial colonies directly informs product release decisions, regulatory compliance, and vaccine potency verification under Good Practice (GxP) and FDA 21 CFR Part 11 requirements. Errors in CFU detection can compromise sterility assurance and delay batch disposition, posing substantial risks to patient safety and supply chain reliability~\cite{who2019}.

Traditionally, CFU counting has relied on manual visual inspection of Petri dishes by trained microbiologists. While effective at a small scale, this process is time-consuming, subjective, and prone to inter-operator variability~\cite{alexander1958,chiang2015}. Ambiguous colony boundaries, overlapping growth, condensation artifacts, and non-uniform illumination further degrade reproducibility. As vaccine manufacturing volumes expand, manual counting has become a significant bottleneck in digital Quality Control (QC) workflows. The transition toward Industry 4.0 manufacturing has therefore created an urgent demand for automated, traceable, and auditable CFU quantification systems~\cite{torelli2018,huang2020}.

Recent advances in deep learning (DL) and computer vision have demonstrated the feasibility of automating colony counting with near-human accuracy. Early works leveraged convolutional neural networks (CNNs) to segment and classify colonies, outperforming traditional morphological methods~\cite{ferrari2017,andreini2018}. Beznik et al. (2020)~\cite{beznik2020} introduced a CNN-based segmentation framework for bacterial and fungal colony detection in vaccine production. Their model significantly improved detection precision but remained sensitive to variations in lighting, agar texture, and colony overlap---conditions common in real manufacturing environments.

Parallel research explored one-stage object detectors for microbial analysis. Whipp and Dong (2022)~\cite{whipp2022} presented a YOLO-based deep learning approach for automated bacterial colony counting using the AGAR dataset. Their models achieved mean Average Precision (mAP@0.5) between 96\% and 99\% for \textit{S. aureus} colonies with inference times below 10 ms per image, confirming YOLO's potential for real-time colony detection. However, the study also noted degraded accuracy on small, overlapping, or vague colonies---a critical limitation for pharmaceutical-grade applications.

Despite these advances, deploying deep learning for regulated QC introduces several persistent challenges:
\begin{enumerate}[itemsep=0pt,parsep=2pt]
    \item \textbf{Domain variability} across manufacturing sites causes performance drift due to changes in lighting, agar color, and incubation conditions~\cite{whipp2022}.
    \item \textbf{Small-object detection} remains difficult for colonies exhibiting partial occlusion or aggregation~\cite{minaee2021}.
    \item \textbf{Explainability and auditability} are limited in conventional DL pipelines, hindering FDA or EMA validation~\cite{wang2023}.
    \item \textbf{Data drift} and evolving microbial morphology require continuous learning and retraining.
    \item \textbf{Integration with enterprise systems} (e.g., SAP-QM) requires deterministic outputs, metadata traceability, and version control for every inference.
\end{enumerate}

To overcome these barriers, this study presents a \textbf{multi-agent framework} that combines object-detection models with semantic interpretation from a vision-language model for robust CFU quantification. Building on Beznik et al. (2020)~\cite{beznik2020} and Whipp \& Dong (2022)~\cite{whipp2022}, we integrate a customized \textbf{Meta Detectron2} model optimized for small-object detection with a VLM-driven decision-validation layer to form an adaptive, explainable, and regulation-ready system. The multi-agent architecture introduces \textit{agentic reasoning}, where DL and VLM agents operate independently, cross-validate colony counts, and escalate discordant cases to human experts for review and retraining.

The proposed framework was deployed within \textbf{GSK's Databricks environment}, leveraging Delta Lake for data versioning, MLflow for experiment tracking, and REST-based integration with Postgres and SAP for automated QC data ingestion. Experimental validation on 50,000 gold standard Petri-dish images demonstrated a \textbf{99\% detection rate}, \textbf{2\% false positives}, and \textbf{0.6\% false negatives}, achieving an \textbf{85\% reduction in human verification workload} and leading to significant cost savings across vaccine sites.

In summary, the primary contributions of this paper are as follows:
\begin{enumerate}[itemsep=0pt,parsep=2pt]
    \item \textbf{Customized Detectron2-Based Detector:} A pharmaceutical-grade CFU detection model optimized for multi-scale small-object sensitivity and image-quality variance.
    \item \textbf{Hybrid Multi-Agent CFU Detection Architecture:} Integration of a DL-based object detection model and a VLM-based validation agent to improve robustness and interpretability in CFU quantification, with a feedback mechanism linking human validation with automated deep learning retraining for adaptive performance in GxP environments.
    \item \textbf{Industrial-Scale MLOps Deployment:} End-to-end implementation within GSK's Databricks ecosystem, demonstrating scalable, auditable, and regulation-ready microbiological quality control.
\end{enumerate}

\section{Related Work}

\subsection{CFU Detection and Counting in Microbiology}

Early efforts to automate CFU enumeration focused on hardware-based counters that employed optical scanning or phototube detection to measure transmitted light across Petri dishes~\cite{mansberg1957,alexander1958}. While these systems provided basic automation, they required manual calibration for each plate and were prone to artifacts such as air bubbles, reflections, and surface irregularities.

The advent of digital imaging and software-based methods enabled thresholding and contour-based segmentation for CFU quantification~\cite{chiang2015,torelli2018}. Although such techniques reduced human effort, they struggled in the presence of overlapping colonies, heterogeneous agar surfaces, or condensation artifacts.

With the rise of \textbf{machine learning (ML)}, early classifiers such as Random Forests demonstrated moderate success in colony detection~\cite{flaccavento2011}. However, \textbf{deep learning (DL)} models quickly became the standard. Ferrari et al. (2017)~\cite{ferrari2017} and Andreini et al. (2018)~\cite{andreini2018} established that convolutional neural networks (CNNs) could significantly outperform classical image processing pipelines in CFU segmentation and classification. Building on this progress, Beznik et al. (2020)~\cite{beznik2020} applied a U-Net architecture with a ResNet encoder to CFU images, achieving robust segmentation and the ability to differentiate virulent and avirulent colonies. Despite these gains, purely CNN-based systems remained vulnerable to low-quality or high-density plates, where overlapping colonies and artifacts led to inconsistent counts and continued reliance on human validation.

\subsection{Deep Learning in Industrial Biopharma Contexts}

Within regulated biomanufacturing environments, deep learning has been integrated into \textbf{industrial-scale microbiological quality control}. GSK, for example, implemented a DL-based binary classifier in its CFU imaging pipeline, achieving over 95\% accuracy and near 100\% sensitivity on negative plates while maintaining human verification for positive or low-quality samples~\cite{beznik2020}. This hybrid ``DL + human'' workflow reduced manual counting efforts by approximately 50\% and led to significant annual operational savings. However, as with other CNN-based systems, model performance degraded under poor lighting or in the presence of dense microbial growth.

To overcome these limitations, several researchers explored \textbf{real-time object detection architectures}. Whipp and Dong (2022)~\cite{whipp2022} presented one of the first YOLO-based frameworks for microbial colony detection, achieving a mean Average Precision (mAP@0.5) between 96\% and 99\%. Their study highlighted YOLO's efficiency for real-time CFU detection but also noted diminished performance on small or overlapping colonies. Subsequent experiments comparing YOLOv5, YOLOv8 within GSK's internal dataset confirmed similar trends, prompting the transition to a \textbf{Meta Detectron2-based detector} with improved feature resolution and multi-scale learning for pharmaceutical-grade robustness.

\subsection{Vision-Language Models for Biological Image Analysis}

Recent breakthroughs in \textbf{multimodal artificial intelligence} have introduced \textbf{vision-language models (VLMs)} capable of jointly processing images and text~\cite{radford2021,alayrac2022}. State-of-the-art architectures such as CLIP, Flamingo, and GPT models have demonstrated strong reasoning capabilities across biomedical imaging and microscopy tasks. By combining visual encoders with language transformers, these systems offer explainable predictions through natural language descriptions---an asset for regulated QC workflows~\cite{chen2023,wang2023}.

Unlike traditional CNNs that output static labels, VLMs can simultaneously perform classification, reasoning, and visual verification. Early benchmarks on CFU datasets have demonstrated that VLMs can differentiate valid vs invalid plates, providing contextual validation before automated counting. Moreover, \textbf{quantized VLM implementations} can now be deployed on enterprise-grade GPUs with minimal performance loss, enabling their adoption in GxP environments.

\subsection{Agentic AI Frameworks and Multi-Agent Orchestration}

In parallel, \textbf{multi-agent frameworks} have emerged as orchestration systems that coordinate multiple specialized AI agents within a unified workflow~\cite{lin2017,singh2024}. A triage agent delegates tasks to sub-agents that are responsible for classification, verification, or feedback integration, ensuring decision reliability and auditability.

In the context of CFU detection, this paradigm has evolved into an \textbf{agentic decision pipeline}. A VLM agent first classifies each plate as valid or invalid. For valid plates, both the DL-based detector (Detectron2) and a VLM independently generate colony counts. If their predictions reconcile within a pre-defined threshold, the results are automatically logged into Postgres and an SAP Quality Management (QM) system, ensuring traceability. If discrepancies arise, the case is escalated to a human expert, whose review is fed back into the system for deep learning retraining and continuous improvement.

This \textbf{consensus-based, self-improving architecture} advances beyond traditional ``AI reviewer'' workflows by embedding reconciliation, escalation, and continuous learning within the same loop. The proposed \textbf{multi-agent framework} demonstrates this approach for CFU detection in vaccine manufacturing, achieving explainability, scalability, and regulatory compliance.

\section{Proposed Methodology}

\subsection{Problem Formulation}

The central objective of this work is to design an \textbf{automated, explainable, and regulation-ready system} for detecting and counting colony-forming units (CFUs) from Petri-dish images used in vaccine and pharmaceutical manufacturing. The system must satisfy two stringent requirements:
\begin{enumerate}[itemsep=0pt,parsep=2pt]
    \item \textbf{Safety assurance:} Ensure zero missed colonies to prevent contaminated batches from being released.
    \item \textbf{Operational efficiency:} Automate high-confidence cases while keeping total inference latency below \textbf{10 seconds per image}.
\end{enumerate}

Formally, let $\mathcal{I} = \{x_1, x_2, \ldots, x_n\}$ denote a set of Petri-dish images. For each image $x_i$, the system estimates the colony count $C(x_i)$ and classification $K(x_i) \in \{\text{bacteria}, \text{mold}\}$. The predictive model $f_\theta$ minimizes:
\begin{equation}
\mathcal{L} = \alpha \mathcal{L}_{\text{SmoothL1}}(C, \hat{C}) + \beta \mathcal{L}_{\text{CE}}(K, \hat{K})
\end{equation}
where $\mathcal{L}_{\text{SmoothL1}}$ is the Smooth L1 loss for counting, and $\mathcal{L}_{\text{CE}}$ is the categorical cross-entropy loss for colony-type classification.

To ensure reliability in production environments, a consensus constraint is introduced between the Deep Learning (DL) detector $f_{\text{DL}}$ and the Vision-Language Model (VLM) $f_{\text{VLM}}$:
\begin{equation}
|f_{\text{DL}}(x_i) - f_{\text{VLM}}(x_i)| < \delta
\end{equation}
where $\delta = 0.05$ represents a 5\% tolerance. Predictions meeting this condition are auto-approved; otherwise, they are escalated for human review. This multi-agent consensus formulation replaces conventional single-model inference with agentic orchestration, ensuring both accuracy and regulatory compliance.

\subsection{Architecture Overview}

\begin{figure}[t]
\centering
\includegraphics[width=\columnwidth]{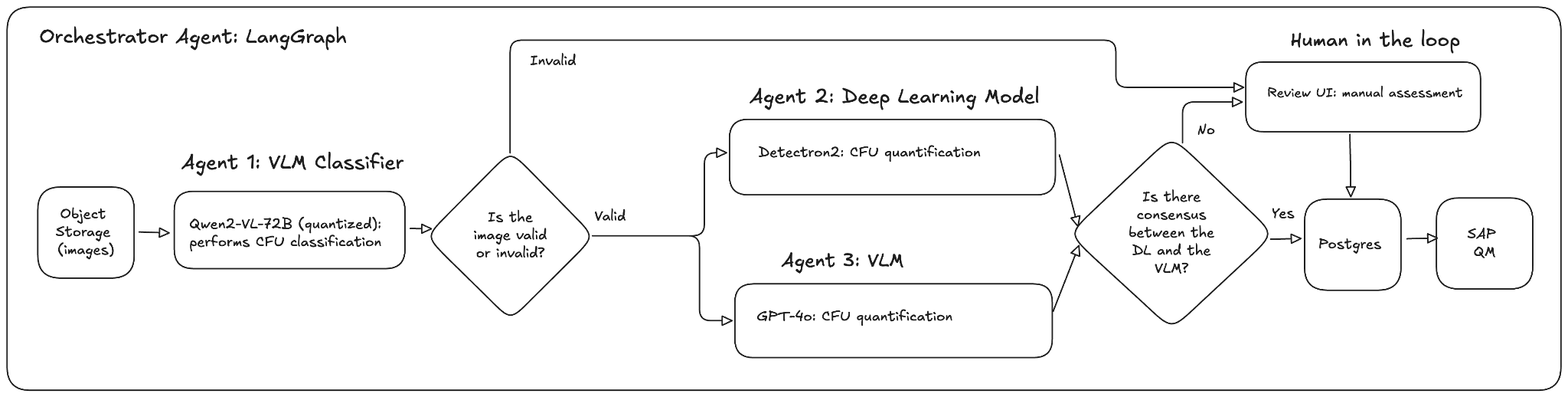}
\caption{High-level architecture diagram of the multi-agent CFU detection system.}
\label{fig:architecture}
\end{figure}

The proposed system employs a \textbf{multi-agent architecture} consisting of three principal components (Figure~\ref{fig:architecture}):

\textbf{1) Vision-Language Model (VLM) Pre-Screener:} The VLM acts as the entry point, classifying Petri-dish images as valid (clear) or invalid (artifactual, contaminated, or poorly focused). The selected model (\textbf{Qwen2-VL-quantized}) balances interpretability and computational efficiency. Invalid images are directly sent to a human reviewer.

\textbf{2) Dual Colony Counting (Detectron2 + GPT-4o):} For valid images, both the deep-learning detector and GPT-4o independently estimate CFU counts. The \textbf{Detectron2-based DL detector} ensures robust small-object detection, while the \textbf{VLM counter} performs semantic reasoning to validate or challenge the DL output.

\textbf{3) Agentic Orchestration Layer:} Implemented using the \textbf{LangGraph framework}, this layer coordinates agent communication, resolves discrepancies, and ensures validated results are automatically published to \textbf{Postgres} and synchronized with \textbf{SAP QM} for regulatory traceability.

When DL and VLM outputs reconcile, results are sent directly to Postgres and SAP QM. Divergent predictions trigger human validation, and expert corrections are logged for continuous retraining and drift prevention.

\subsection{Deep Learning Detector}

The \textbf{Deep Learning (DL) detector} is built on \textbf{Detectron2} with a ResNet-101 + FPN backbone for multi-scale feature extraction and colony localization. It was initially trained on \textbf{10,000 high-quality Petri-dish images}, followed by retraining with \textbf{40,000 production samples}, enabling strong domain generalization.

\subsubsection{Technical Optimizations}

To achieve near-human accuracy (99\% detection rate), several key optimizations were applied:
\begin{itemize}[itemsep=0pt,parsep=2pt]
    \item \textbf{Backbone:} ResNet-101 + Feature Pyramid Network (FPN) for multi-scale detection.
    \item \textbf{Anchor Tuning:} Custom anchor scales (8, 16, 32, 64) and aspect ratios (0.5, 1.0, 2.0) optimized for 5--30 px colonies.
    \item \textbf{Augmentation:} Adaptive brightness, contrast normalization, Gaussian blur, and elastic deformation to simulate environmental variability.
    \item \textbf{Loss Functions:} Hybrid \textbf{Focal + Smooth L1} loss for class imbalance and localization precision.
    \item \textbf{Learning Rate Policy:} Cyclical LR with cosine annealing (base LR = $2.5 \times 10^{-4}$).
    \item \textbf{Soft-NMS:} IoU = 0.4 to retain overlapping colonies.
    \item \textbf{Mixed Precision:} FP16 training on Databricks clusters for accelerated convergence and reduced memory use.
\end{itemize}

The optimized model achieved \textbf{mAP@0.5 = 99.0\%}, \textbf{F1 = 0.985}, outperforming YOLOv7/YOLOv8 by approximately 3--4\% in both precision and recall.

\subsubsection{CNN Model Selection for Bacteria-Mold Classification}

Each colony detected by Detectron2 is further classified as \textbf{bacterial} or \textbf{mold} using one of six candidate CNN architectures: \textbf{Inception}, \textbf{ResNet101}, \textbf{DenseNet121}, \textbf{Xception}, \textbf{VGG16}, and \textbf{ResNet50}.

Instead of soft-voting, a \textbf{dynamic model-selection mechanism} is used:
\begin{itemize}[itemsep=0pt,parsep=2pt]
    \item Each CNN is trained and validated via 5-fold cross-validation.
    \item Evaluation metrics: \textbf{Balanced F1}, \textbf{Recall}, \textbf{ROC-AUC}, \textbf{Calibration Error}, and \textbf{Latency}.
    \item The \textbf{top-performing model} is automatically promoted to production, tracked in \textbf{MLflow}, and logged in \textbf{PostgreSQL} before storage in SAP QM.
    \item \textbf{Live performance} is continuously monitored and triggers retraining or replacement upon degradation.
\end{itemize}

This adaptive selection ensures consistent accuracy under domain shift.

\subsection{Vision-Language Model (VLM) Pre-Screener and Counter}

The \textbf{VLM model} serves both as a \textbf{pre-screener} and an \textbf{independent reviewer}.

\subsubsection{Pre-Screening Stage}

The VLM (Qwen2-VL-quantized model) filters out invalid images (condensation, glare, blur) before DL inference, reducing compute load and preventing artifactual detections. Quantization (INT4/INT8 via GPTQ) cuts GPU memory by approximately 75\% with less than 1\% accuracy loss.

\subsubsection{Counting Stage}

The VLM (GPT-4o) performs zero-shot CFU counting via structured visual-text reasoning. The model outputs structured JSON:
\begin{verbatim}
{"plate_id": id, "quality": "valid/invalid",
 "count": n, "reason": text}
\end{verbatim}
These outputs feed directly into \textbf{Postgres} and \textbf{SAP QM}.

\textbf{Inference time:} 5--7 s per plate on 1$\times$ NVIDIA A100 GPU, within the 10s SLA.

\subsection{Agentic Orchestration Layer}

The \textbf{Agentic Orchestration Layer}, implemented using \textbf{LangGraph}, governs inter-agent communication and decision reconciliation:
\begin{itemize}[itemsep=0pt,parsep=2pt]
    \item \textbf{Agent 1 (VLM Classifier):} Qwen2-VL-quantized performs valid/invalid classification.
    \item \textbf{Agent 2 (Deep Learning Model):} Detectron2 + CNN performs CFU count.
    \item \textbf{Agent 3 (VLM Counter):} GPT-4o performs CFU count.
    \item \textbf{Human in the Loop:} Escalation to human reviewer when required, with feedback loop.
\end{itemize}

\textbf{Workflow:}
\begin{enumerate}[itemsep=0pt,parsep=2pt]
    \item Agent 1 filters invalid plates.
    \item Agents 2 and 3 perform independent CFU counts.
    \item If $\Delta\text{CFU} \leq 5\%$, results auto-approve $\rightarrow$ stored in \textbf{Postgres} $\rightarrow$ sent to \textbf{SAP QM}.
    \item Otherwise, cases escalate to Human Reviewer.
    \item All interactions are logged and trigger Databricks deep learning retraining using validated feedback.
\end{enumerate}

This modular design supports asynchronous operation and ensures $<$10 s end-to-end latency.

\subsection{Computational Efficiency}

\begin{itemize}[itemsep=0pt,parsep=2pt]
    \item \textbf{DL Inference:} 3--5 s/plate (GPU).
    \item \textbf{VLM Inference:} 6--7 s/plate (quantized 72B).
    \item \textbf{End-to-End Pipeline:} $<$10 s/plate (production SLA).
    \item \textbf{Bad Plate Bypass:} $\sim$40\% compute savings.
\end{itemize}

Quantization enables the model to be compressed so it can run on a smaller compute cluster, thereby reducing long-term serving costs.

\subsection{Implementation and Deployment}

\begin{itemize}[itemsep=0pt,parsep=2pt]
    \item \textbf{Ecosystem:} Detectron2, Qwen2-VL-quantized, GPT-4o, LangGraph, MLflow (Databricks).
    \item \textbf{Compute:} Databricks GPU clusters (A100/V100); CPU fallback for negative plates.
    \item \textbf{Data Storage:} Images in \textbf{Azure Blob Storage}, metadata in \textbf{Postgres}.
    \item \textbf{Integration:} Outputs pushed to \textbf{SAP QM} for regulatory traceability.
    \item \textbf{Security:} AAD RBAC, HTTPS encryption, and Azure Key Vault secrets.
\end{itemize}

This setup guarantees scalability, auditability, and compliance with GxP and FDA 21 CFR Part 11 standards.

\section{Results and Observations}

\subsection{Quantitative Evaluation}

The performance of the proposed multi-agent CFU detection framework was quantitatively evaluated on a dataset comprising over 50,000 Petri-dish images. The Detectron2-based deep learning (DL) detector achieved a mean Average Precision (mAP@0.5) of \textbf{99.0\%}, with \textbf{precision = 98.8\%} and \textbf{recall = 98.5\%}. The false-positive rate (FPR) and false-negative rate (FNR) were \textbf{2.0\%} and \textbf{0.6\%}, respectively. These results exceed the performance of YOLOv7 and YOLOv8 baselines by approximately \textbf{3--4\%}, establishing Detectron2 as the most effective object detection backbone for pharmaceutical-grade colony detection and enumeration.

A comparative evaluation across five architectures---YOLOv5, YOLOv7, YOLOv8, Mask R-CNN, and Detectron2---demonstrated the superiority of the proposed model in both accuracy and robustness under heterogeneous image conditions. The quantitative comparison is presented in Table~\ref{tab:dl_comparison}.

\begin{table}[t]
\centering
\caption{Performance Comparison of Deep Learning Models for CFU Detection}
\label{tab:dl_comparison}
\begin{tabular}{@{}lccccc@{}}
\toprule
\textbf{Model} & \textbf{mAP} & \textbf{Prec.} & \textbf{Rec.} & \textbf{FNR} & \textbf{FPR} \\
\midrule
YOLOv5 & 96.2 & 96.0 & 95.8 & 2.9 & 3.5 \\
YOLOv7 & 96.8 & 97.0 & 96.1 & 2.5 & 3.1 \\
YOLOv8 & 97.1 & 97.2 & 96.5 & 2.3 & 2.9 \\
Mask R-CNN & 98.2 & 98.1 & 97.4 & 1.8 & 2.3 \\
\textbf{Detectron2} & \textbf{99.0} & \textbf{98.8} & \textbf{98.5} & \textbf{0.6} & \textbf{2.0} \\
\bottomrule
\end{tabular}
\end{table}

\subsection{Vision Language Model (VLM) Performance}

A two-stage evaluation framework was developed to assess the effectiveness of vision-language models (VLMs) in automating Petri-dish quality assessment. The first stage focused on distinguishing between \textbf{valid and invalid plates}, while the second stage evaluated \textbf{colony-forming unit (CFU) count estimation} using multiple VLM architectures.

\subsubsection{Invalid Plate Detection}

To ensure high-quality inputs for downstream CFU quantification, a VLM-based pre-screening module was developed to classify Petri-dish images as \textit{valid} or \textit{invalid} automatically. Five state-of-the-art vision-language models---GPT-4o, Pixtral, Qwen2-VL-Quantized, Qwen2-VL-7B, and Qwen-VL-72B---were benchmarked on a dataset of \textbf{451 samples}. Performance was assessed using the False Negative Rate (FNR), Detection Rate (DR), False Positive Rate (FPR), and Invalid Plate Detection Rate (NPDR). Table~\ref{tab:invalid_detection} presents the comparative results.

\begin{table}[t]
\centering
\caption{Invalid Plate Detection Performance (N=451)}
\label{tab:invalid_detection}
\begin{tabular}{@{}lcccc@{}}
\toprule
\textbf{Model} & \textbf{FNR} & \textbf{DR} & \textbf{FPR} & \textbf{NPDR} \\
\midrule
GPT-4o & 0.05 & 0.95 & 0.30 & 0.70 \\
Pixtral & 0.23 & 0.77 & 0.01 & 0.99 \\
Qwen2-Quant. & 0.24 & 0.76 & 0.01 & 0.99 \\
Qwen2-VL-7B & 0.14 & 0.86 & 0.08 & 0.92 \\
Qwen-VL-72B & 0.23 & 0.77 & 0.00 & 1.00 \\
\bottomrule
\end{tabular}
\end{table}

GPT-4o achieved the strongest sensitivity (lowest FNR = 0.05), indicating reliable detection of valid plates. However, its relatively high FPR (0.30) implies a substantial number of valid images being incorrectly rejected. Pixtral, Qwen2-Quantized, and Qwen-VL-72B demonstrated \textbf{exceptionally low FPR ($\leq$0.01)} and high NPDR (0.99--1.00), making them more robust for filtering invalid samples. Among these, \textbf{Qwen2-Quantized} offered the most favorable operational balance, with high accuracy and significantly lower inference latency, making it suitable for real-time deployment.

\subsubsection{CFU Count Detection and Model Benchmarking}

Following pre-screening, CFU quantification was performed using five multimodal VLMs evaluated on \textbf{1,152 production-grade Petri-dish images}. Predicted colony counts were compared with reference CFU values, with mismatches adjudicated through human expert review. Table~\ref{tab:cfu_validation} summarizes the comparative performance.

\begin{table}[t]
\centering
\caption{CFU Count Validation Across VLM Models (N=1,152)}
\label{tab:cfu_validation}
\begin{tabular}{@{}lccccc@{}}
\toprule
\textbf{Model} & \textbf{Match} & \textbf{Mismatch} & \textbf{Appr.} & \textbf{Verify} \\
\midrule
GPT-4o & 800 & 573 & 69\% & 50\% \\
Pixtral & 313 & 839 & 27\% & 73\% \\
Qwen2-Quant. & 317 & 835 & 28\% & 72\% \\
Qwen2-VL-7B & 297 & 854 & 26\% & 74\% \\
Qwen-VL-72B & 308 & 843 & 27\% & 73\% \\
\bottomrule
\end{tabular}
\end{table}

GPT-4o achieved the highest agreement with ground-truth CFU values, attaining an approval rate of \textbf{69\%}, which is approximately \textbf{2.5$\times$ higher} than other models. It also produced the lowest ``to-verify'' rate (50\%), reducing expert workload significantly. The superior performance of GPT-4o reflects its strong multimodal reasoning and resilience to challenging imaging conditions.

\subsection{Integrated System Performance}

The integration of the Detectron2 detector with LangGraph-based agentic orchestration yielded the following overall system performance:
\begin{itemize}[itemsep=0pt,parsep=2pt]
    \item \textbf{Detection Accuracy:} 99.0\%
    \item \textbf{False Negatives:} 0.6\%
    \item \textbf{False Positives:} 2.0\%
    \item \textbf{Human Verification Reduction:} 85\%
    \item \textbf{Average Inference Latency:} $<$10 seconds per plate
\end{itemize}

The dual-agent consensus mechanism ($\Delta$CFU $\leq$ 5\%) automatically approved \textbf{85\%} of samples, with only \textbf{15\%} requiring manual review. Feedback from verified cases was continuously leveraged within Databricks pipelines for automated retraining, leading to progressive improvement in consensus accuracy over time.

\subsection{Qualitative and Operational Observations}

Qualitative assessments demonstrated that the system remained robust across diverse real-world imaging conditions, effectively managing \textbf{partially occluded colonies}, \textbf{heterogeneous agar coloration}, and \textbf{glare-induced noise}. The VLM components consistently produced interpretable, natural-language explanations for their decisions, enhancing traceability and supporting regulatory requirements for explainable AI in microbiological workflows.

Operational evaluations further highlighted the stability of the multi-agent architecture. The \textbf{agentic orchestration layer} ensured deterministic data routing, end-to-end traceability within the \textbf{SAP QM} environment, and automated audit log generation for compliance monitoring. The combination of \textbf{Detectron2's high-precision visual localization} and \textbf{GPT-4o's advanced multimodal reasoning} enabled accurate and reliable CFU quantification, even under challenging imaging conditions.

Collectively, these observations confirm that the proposed multi-agent architecture transitions CFU assessment from a \textbf{human-in-the-loop} to a \textbf{human-on-the-loop} paradigm, establishing a new benchmark for \textbf{autonomous, explainable, and regulation-compliant AI} in vaccine manufacturing and pharmaceutical quality control.

\section{Discussion}

The results demonstrate that integrating VLMs into a multi-agent analytical architecture can substantially improve the reliability, traceability, and scalability of CFU quantification workflows in pharmaceutical quality control. The comparative evaluation shows that no single model excels uniformly across all tasks; instead, distinct VLMs offer complementary strengths. For instance, GPT-4o exhibited superior multimodal reasoning for CFU enumeration, while Qwen2-Quantized provided optimal performance for invalid-plate screening at significantly lower computational cost.

This separation of responsibilities highlights the importance of task-specific VLM allocation within automated microbiological analysis pipelines. The use of a lightweight, high-precision invalid-plate filter (Qwen2-Quantized) ensured that only high-quality inputs were forwarded to GPT-4o for downstream CFU estimation, thereby reducing misclassification propagation and improving end-to-end accuracy.

The deployment of an agentic orchestration layer further strengthened system resilience. Deterministic data routing, SAP QM integration, and automated audit logging contributed to full traceability---an essential requirement in regulated environments. Moreover, VLM-generated natural-language rationales enhanced interpretability, supporting human oversight and compliance with emerging guidelines for explainable AI in manufacturing.

Overall, the findings underscore the value of combining classical vision models (e.g., Detectron2) with modern VLMs to build robust, interpretable, and regulation-compliant analytical systems capable of transitioning from human-intensive inspection toward autonomous microbiological analysis.

\section{Conclusion}

This study presents a multi-agent, VLM and DL-driven architecture for autonomous CFU quantification that significantly improves accuracy, reliability, and operational efficiency in vaccine and biopharmaceutical manufacturing. Through systematic benchmarking, \textbf{Qwen2-VL-Quantized} was identified as the most effective model for invalid-plate pre-screening, while \textbf{GPT-4o} demonstrated superior concordance with ground-truth CFU counts. The combined use of the \textbf{Detectron2 vision model} and the \textbf{GPT-4o multimodal model} created a more accurate and stable analytical pipeline. Together, they reduced error propagation, increased the rate of automatic approvals, and lowered the need for expert review.

From an operational standpoint, the dual-agent DL and GPT-4o consensus mechanism reduced human verification requirements by 85\% while maintaining inference latency under 10 seconds per plate. These efficiencies translate into significant annual cost savings, underscoring the tangible economic value of the proposed architecture. The integration of an automated, self-improving retraining loop within Databricks further ensures continuous performance refinement driven by real production data.

The agentic orchestration layer provided deterministic data routing, end-to-end traceability within the \textbf{SAP QM} system, and automated audit log generation---key enablers for regulatory compliance and transparent decision-making. The system's ability to generate interpretable, natural-language rationales additionally supports explainability mandates for AI in regulated environments.

Overall, the proposed architecture marks a paradigm shift from \textbf{human-in-the-loop} to \textbf{human-on-the-loop} microbiological quality control, establishing a scalable, explainable, and regulation-ready framework for automated microbial colony counting. Future work will extend this architecture toward \textbf{multi-species detection}, \textbf{adaptive thresholding}, and \textbf{edge deployment}, enabling real-time, smart-manufacturing applications across broader pharmaceutical and bioprocessing domains.

\bibliographystyle{ieeetr}

\end{document}